\documentclass[a4paper, 10pt, conference]{ieeeconf}      

\IEEEoverridecommandlockouts                              

\usepackage{graphicx} 
\usepackage{subfig}
\usepackage{amsmath} 
\usepackage{algorithmic}
\usepackage{siunitx}
\usepackage{gensymb}
\usepackage{color}
\usepackage{fancyref}
\let\DeclareUSUnit\DeclareSIUnit
\let\US\SI
\DeclareUSUnit\inch{in}
\newcommand*{\fancyreflstlabelprefix}{alg}

\fancyrefaddcaptions{english}{%
  \providecommand*{\freflstname}{algorithm}%
  \providecommand*{\Freflstname}{Algorithm}%
}

\frefformat{plain}{\fancyreflstlabelprefix}{\freflstname\fancyrefdefaultspacing#1}
\Frefformat{plain}{\fancyreflstlabelprefix}{\Freflstname\fancyrefdefaultspacing#1}

\frefformat{vario}{\fancyreflstlabelprefix}{%
  \freflstname\fancyrefdefaultspacing#1#3%
}
\Frefformat{vario}{\fancyreflstlabelprefix}{%
  \Freflstname\fancyrefdefaultspacing#1#3%
}
\usepackage{color}
\usepackage{tikz}

\definecolor{floor}{RGB}{228,26,28}
\definecolor{wall}{RGB}{55,126,184}
\definecolor{ceiling}{RGB}{255,127,0}
\definecolor{table}{RGB}{77,175,74}
\definecolor{chair}{RGB}{247,129,191}
\definecolor{cabinet}{RGB}{255,255,51}
\definecolor{object}{RGB}{152,78,163}
\definecolor{unknown}{RGB}{166,86,40}

\usepackage{multirow}
\usepackage{booktabs}

\newsavebox{\ieeealgbox}
\newenvironment{boxedalgorithmic}
  {\begin{lrbox}{\ieeealgbox}
   \begin{minipage}{\dimexpr\columnwidth-2\fboxsep-2\fboxrule}
   \begin{algorithmic}}
  {\end{algorithmic}
   \end{minipage}
   \end{lrbox}\noindent\fbox{\usebox{\ieeealgbox}}}

\makeatletter
\let\saved@reftextfaraway\reftextfaraway
\renewcommand*{\reftextfaraway}[1]{%
  \begingroup
    \def\ref@unknown@value{??}%
    \ifx\@tempa\ref@unknown@value
      \count@=0 %
    \else
      \count@\thevpagerefnum\relax
      \advance\count@ by -\@tempa\relax
      \ifnum\count@<0 \count@=-\count@\fi
    \fi
    \ifnum\count@<5 %
      \unskip
    \else
      \saved@reftextfaraway{#1}%
    \fi
  \endgroup
}
\makeatother

\title{\LARGE \bf
Where to look first? Behaviour control for fetch-and-carry missions of service robots
}

\author{Markus Bajones$^{1}$, Daniel Wolf$^{1}$, Johann Prankl$^{1}$, Markus Vincze$^{1}$
\thanks{$^{1}$Faculty of Electrical Engineering and Information Technology,
		Vienna University of Technology, 1040 Vienna, Austria
		{\tt\small \{bajones, wolf, prankl, vincze\}@acin.tuwien.ac.at}}%
}

\begin{document}

\maketitle
\thispagestyle{empty}
\pagestyle{empty}

\begin{abstract}
This paper presents the behaviour control of a
service robot for intelligent object search in a domestic environment.
A major challenge in service robotics is to enable
fetch-and-carry missions that are satisfying for the user in
terms of efficiency and human-oriented perception.
The proposed behaviour controller provides an informed intelligent
search based on a semantic segmentation
framework for indoor scenes and integrates it with object recognition and grasping.
Instead of manually annotating search positions in
the environment, the framework automatically suggests likely
locations to search for an object based on contextual information, e.g. next to tables and shelves.
In a preliminary set of experiments we demonstrate that
this behaviour control is as efficient as using manually annotated
locations. Moreover, we argue that our approach will reduce
the intensity of labour associated with programming fetch-and-carry
tasks for service robots and that it will be perceived as
more human-oriented.

\end{abstract}

\section{Introduction}
The ability of a robot autonomously interacting with a human requires sophisticated cognitive skills including perception, navigation, decision making and learning. Impressive achievements have already been made in the research field of HRI considering robots as tour guides in museums \cite{Burgard1998, Thrun1999}, shopping malls \cite{Gross2008} and also for assistive robots in the care and domestic context \cite{Graf2009}. 

However, one of the biggest challenges still is the integration of methods into operational autonomous systems for the domestic context, which achieve satisfying results for their end-users. Often a miss-match between user expectations and robot performance can be observed in HRI studies \cite{Lohse2009}, for instance because the robot behaviour is not legible to the users or is simply too slow. Imagine you command your household robot to bring you your mug and the robot could only search for it at pre-programmed places: What if the mug is at none of these places or the user rearranged the flat? Another solution could be that the robot would navigate to the center of the room, rotate several times and then start a time-consuming brute-force object detection everywhere in the map. This behaviour would already increase flexibility, but would still not be very legible for the user: Why getting an overview and then start a time-consuming search?

Would it not be much more intelligent and legible if the robot first gets an overview of the environment and then looked at the most probable locations for the mug to be, e.g. on tables or in the cupboard? One way to gain the information about this kind of relations is to extract it from knowledge databases. 

\begin{figure}[thpb]
\centering
\includegraphics[scale=0.28]{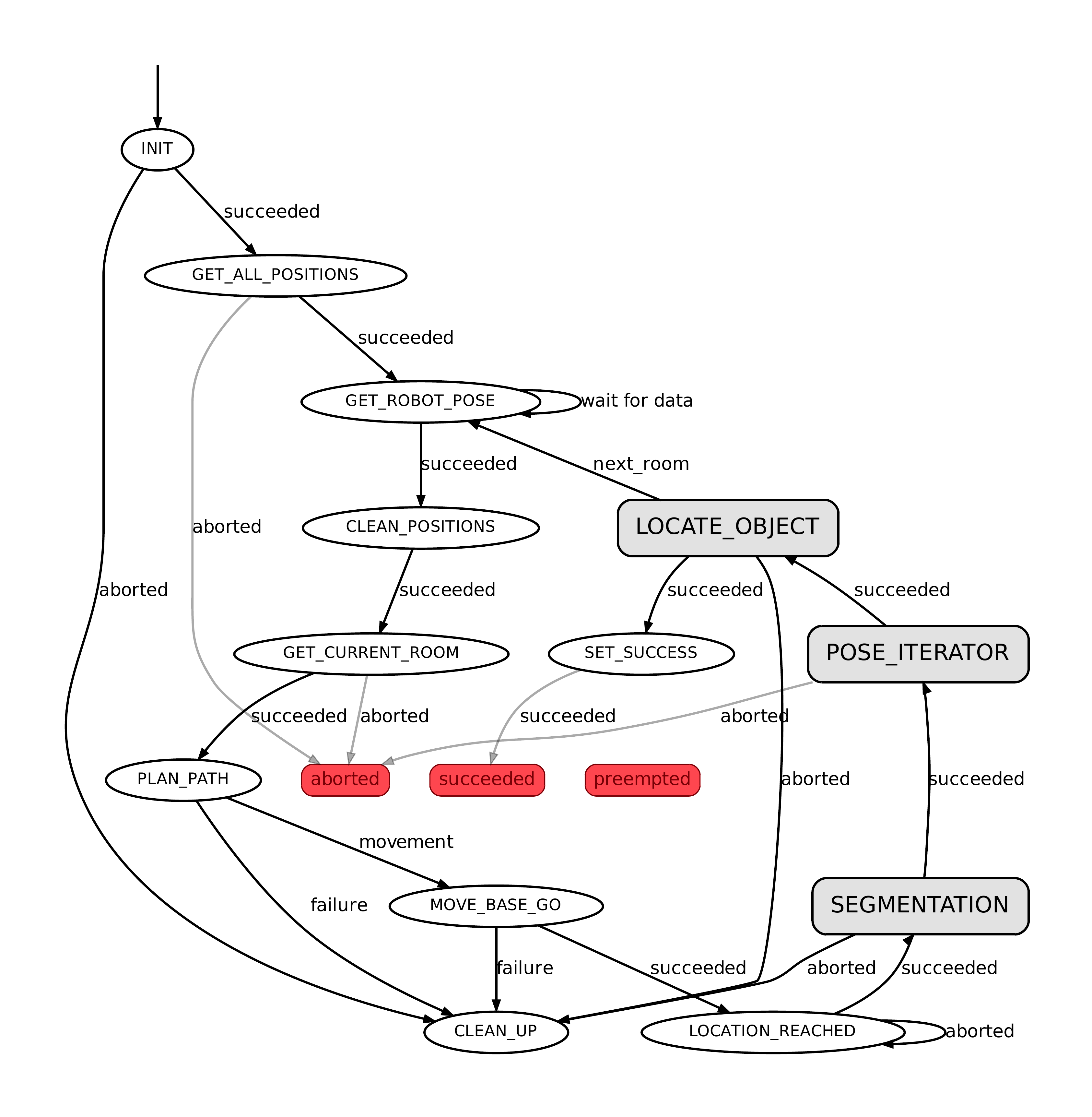}
\caption{Simplified state machine representation of our ``fetch-and-carry'' object task approach. Grey states denote sub-state machines.}
\label{fig:BO_SM}
\end{figure}

Our solution for this problem is to develop a flexible behaviour controller for fetch-and-carry tasks implemented with SMACH, a Python-based library for building hierarchical concurrent state machines, as pictured for our proposed implementation in Fig. \ref{fig:BO_SM}. 
We present a framework that is able to generate locations on-the-fly without the need of pre-learned object-location relationship and show in a first experiment in an ambient-assistive-living lab to proof the robustness of our approach.

The remainder of this paper is structured as follows: First, we review related research on service robots performing fetch-and-carry tasks in Section \ref{sec:rel}. We then outline how the SMACH framework works in Section \ref{sec:smach}. Each of the modules and their integration in the behaviour controller is then described in Section \ref{sec:implementation} with a focus on the modules, but not the individual methods. Finally, a first experiment in a living room setting is presented in Section \ref{sec:exp}. A summary of lessons learned from the experiment and an outlook on future work conclude the paper. 

\section{Related Work}
\label{sec:rel}
Object search and fetch-and-carry tasks represent one of the most common tasks of service robots. In traditional fetch-and-carry tasks the robot "simply" has to pick up an object from a known location and deliver it to the user. For instance, in the RoboCup@Home competition 2008/9~\cite{Nardi2009} this has been one of the major tasks. The complexity of this task can vary, e.g.\ the robot has to identify the correct object from a set of objects or in a cluttered scene or the robot gets a very concrete instruction from where to fetch it, such as ``Bring me the mug from the kitchen table". Thus, fetch-and-carry also includes object recognition and autonomous grasping, popular examples are e.g.\ \cite{Bohren2011} and \cite{Stuckler2012}. 

Recently, the perception part has come more into the focus of fetch-and-carry tasks by adding the search for object components. A popular example is George, the curios attentive semantic robot \cite{meger2008curious}. The idea of George is similar to that presented in this paper: Make a robot's search behaviour more intelligent. An intelligent search procedure is more efficient and therefore the fetch-and-carry task become more satisfying for the end-user. \cite{Huettenrauch2002} and  \cite{Walters2007} have already shown that there is the need to make robot's behaviour more efficient and legible for the user.

Therefore, the aim of the work presented in the following is to enable a domestic service robot to perform fetch-and-carry tasks in a user-satisfying manner. To that end, a cognitive robotic system needs to integrate multiple different subsystems. Different tools, libraries and frameworks to develop robotic architectures already exist, such as RobletRTechnology by Baier et al.\ \cite{Baier2006}, graphical tools such as Choreographe and NAOqi for Nao robots \cite{Pot2009}, the behaviour Markup Language (BML) \cite{Kopp2006} and the Robot Operating system ROS \cite{Quigley2009}. A good overview and discussion on the advantages and disadvantages of all these approaches is given in \cite{Siepman2013}.

In the following we will present SMACH, a Python-based framework that allow ROS module integration, we used to develop a behaviour control combining intelligent object search, grasping objects, and bringing them to the user.

\section{SMACH}
\label{sec:smach}
SMACH is a Python-based library for building hierarchical concurrent state machines, which also provides a ROS integrated module to design and execute simple tasks as well as complex robot behaviours. SMACH provides the possibility to quickly create prototype state machines by reusing Python design patterns. 
Within SMACH a task is defined by the following elements:
\begin{itemize}
\item{\textbf{State}} represents a blocking execution with pre-defined outcomes. The result of a state specifies the transitions to the next states.
\item{\textbf{Container}} is a set of one or multiple states and defines their final outcome based on the integrated states. Especially important and useful are the StateMachine and the Concurrence containers. In the StateMachine container all the states are executed one at a time, whereas in a Concurrence multiple states are executed in parallel. 
\item{\textbf{Transitions}} define, given an outcome, to which state or container the execution chain should turn to.
\item{\textbf{Userdata}} is data which can be shared between multiple states and containers, allowing to adapt the state outcome based on accumulated data from previously executed states.
\end{itemize}
For seamless integration SMACH provides interfaces to the three communication methods available in ROS, i.e.\ messages, services and Actionlib. Provided is the MonitorState for listening to published topics, executing a service represented as a state with ServiceState, calling an Actionlib interface within SimpleActionState as well as the possibility of wrapping a state machine inside an Actionlib server. With SMACH viewer, a tool to debug and visualize the running state machine, the provided user data as well as the currently executed states and containers, is also integrated.

\section{Implementation}
\label{sec:implementation}
\begin{figure}
\includegraphics[width=0.45\textwidth]{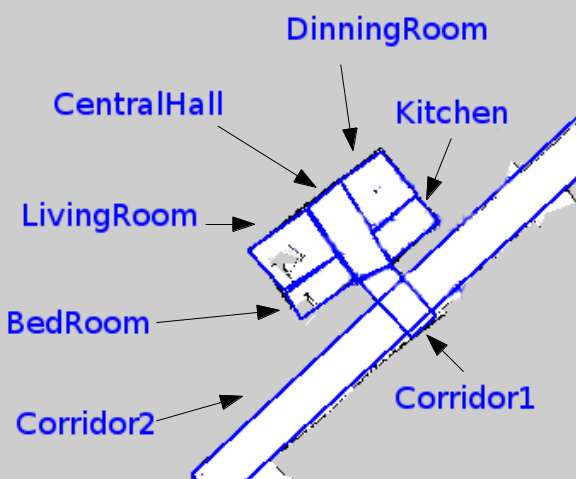}  
\caption{Virtual room arrangement in our ambient-assistive-living lab.}
\label{fig:room_labels}
\end{figure}

For the fetch-and-carry scenario we compare two different implementations within a ROS environment and the SMACH state machine architecture. Both methods rely on a SLAM based 2D map of our laboratory in which we annotate virtual rooms as seen in Fig. \ref{fig:room_labels} to create a multiple room scenario. 

\subsection{Manual definition of search positions}
Starting from the SLAM-based map we manually place ``search positions'' at user-defined locations inside the map. To determine the pose at which the robot begins its task we define a cost function
\begin{equation}
 c(x,y) = c_{bat}(y) - k_1 c_{prob}(x,y) + k_2 c_{pen}(y)
\end{equation}
modelling the cost to search for a given object $x$ at each possible search location $y$. $c_{bat}(y)$ represents the cost of the battery usage estimated by the path length to the pose associated to search location $y$. $c_{prob}(x,y)$ is the probability of object $x$ being found at location $y$ transformed to a cost and is initialized with a uniform distribution across all search locations. The constant factors $k_1$ and $k_2$ are introduced to normalize the cost levels. $c_{pen}(y)$ is a penalty term depending on the room the user is currently in:
\begin{equation}
 c_{pen}(y) = \begin{cases}
               k_{pen} & y \text{ in same room as user} \\
               0 & \text{otherwise}
              \end{cases}              
\end{equation}
The penalty term is based on the assumption that the user does not ask the robot to look for objects which are in the same room as the user himself. It ensures that the robot does not search in this room unless the object has not been detected at the locations in all other possible rooms. 
During the search task all the objects, which can be recognized, and are localized at a certain position trigger the adaptation of $c_{prob}$. Through this simple method Fig. \ref{alg:locate_object} we will learn in course of time where the user places objects and where the robot should start its search. This provides a way to learn the location of specific objects based on the users preference and does not leverage upon a pre-defined knowledge database or data mining from online sources.

\begin{figure}[htpb]
\begin{boxedalgorithmic}
\WHILE{object located is FALSE}
\FORALL{search positions} 
\STATE Calculate path length $c_{obj}$
\IF{available} 
\STATE Look up $c_{prob}$
\STATE Look up search positions room $r_{sp}$
\ENDIF
\IF{$r_{user} \neq r_{sp}$}
\STATE $c_{penalty} = 0$
\ELSE
\STATE $c_{penalty} = 1$
\ENDIF
\IF{$c_{obj} < c_{min}$}
\STATE $c_{min} = c_{obj}$
\ENDIF
\ENDFOR
\STATE Move to search position
\STATE Remove location from list of search positions
\STATE Start object recognition
\IF{object is recognized}
\STATE Set object located to TRUE
\STATE Grasp object
\ENDIF
\IF{Grasp succeeded}
\STATE Put object on tray
\ENDIF
\STATE Update probabilities of all recognized objects
\STATE Move to the user
\STATE Inform the user about the object location
\ENDWHILE 
\end{boxedalgorithmic}
\caption{Algorithm: locate object procedure}
\label{alg:locate_object}
\end{figure}

\subsection{On the fly extraction of search locations}
Our proposed approach automatically generates search positions on the fly after entering a room and executing the algorithm in Fig. \ref{alg:gen_search_poses} which uses semantic segmentation, returning at least 2 search positions or none at all (when no table planes were detected).
\begin{figure}[htpb]
\begin{boxedalgorithmic}
\STATE Move to center pose in a room
\WHILE{rotation $< 360\degree$}
\STATE Rotate $30 \degree$ ccw
\STATE Call semantic segmentation service
\STATE Receive possible search poses
\STATE Remove poses from outside of the known map and outside the current room
\ENDWHILE
\end{boxedalgorithmic}
\caption{Algorithm: obtain search poses}
\label{alg:gen_search_poses}
\end{figure}

\subsubsection*{Semantic segmentation}
\begin{figure}[htbp]
  \centering
  \subfloat[]{\includegraphics[width=0.225\textwidth]{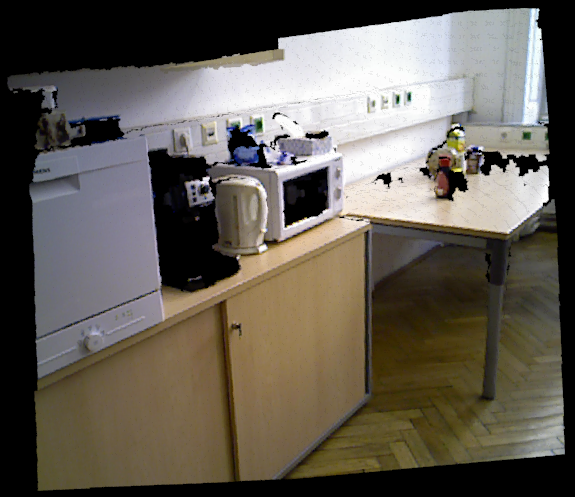}
  \label{fig:pipeline_input}}
  \subfloat[]{\includegraphics[width=0.225\textwidth]{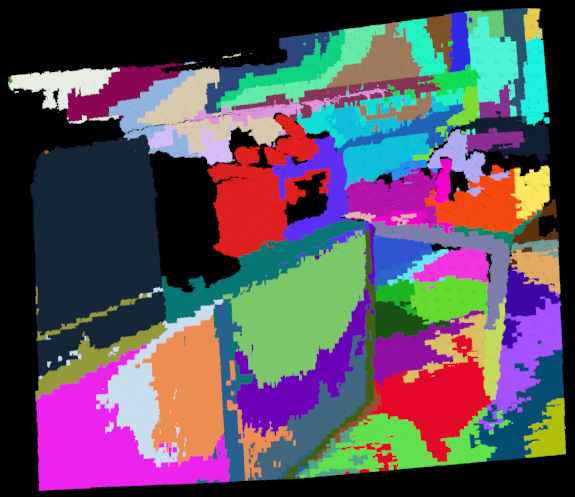}
  \label{fig:pipeline_supervoxels}}\\
  \subfloat[]{\includegraphics[width=0.225\textwidth]{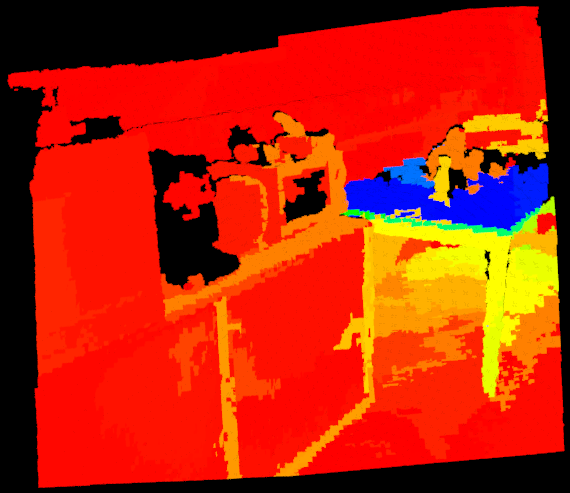} 
    \label{fig:pipeline_classifier}}
  \subfloat[]{\includegraphics[width=0.225\textwidth]{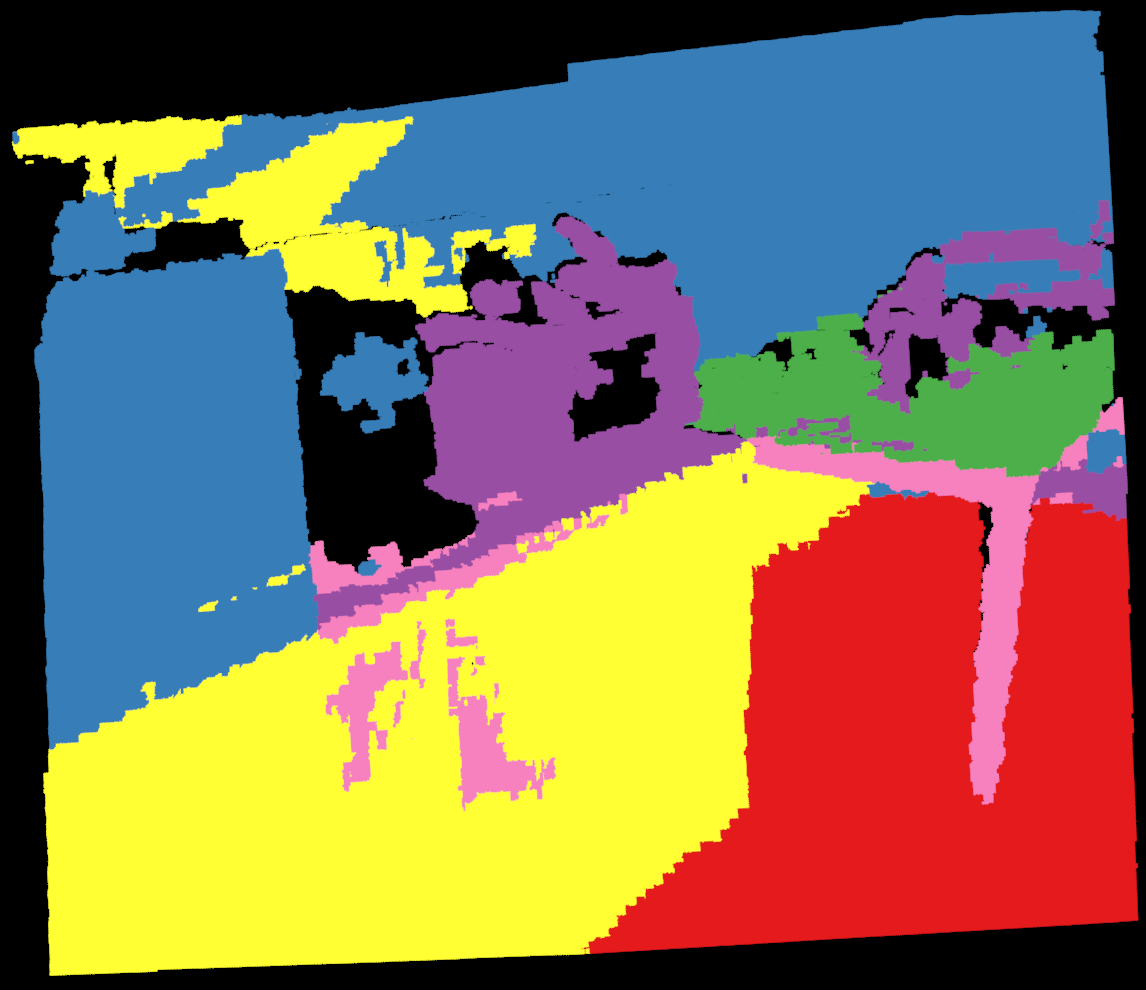}
    \label{fig:pipeline_labeling}}\\
  \subfloat{
  \begin{tikzpicture}\filldraw[draw=black,fill=floor] (1,1) rectangle (1.4,1.25); \end{tikzpicture} floor
  \begin{tikzpicture}\filldraw[draw=black,fill=wall] (1,1) rectangle (1.4,1.25); \end{tikzpicture} wall  
  \begin{tikzpicture}\filldraw[draw=black,fill=ceiling] (1,1) rectangle (1.4,1.25); \end{tikzpicture} ceiling
  \begin{tikzpicture}\filldraw[draw=black,fill=table] (1,1) rectangle (1.4,1.25); \end{tikzpicture} table
  }\\
  \setcounter{subfigure}{4}
  \subfloat[][]{
  \begin{tikzpicture}\filldraw[draw=black,fill=chair] (1,1) rectangle (1.4,1.25); \end{tikzpicture} chair
  \begin{tikzpicture}\filldraw[draw=black,fill=cabinet] (1,1) rectangle (1.4,1.25); \end{tikzpicture} cabinet
  \begin{tikzpicture}\filldraw[draw=black,fill=object] (1,1) rectangle (1.4,1.25); \end{tikzpicture} object
  \begin{tikzpicture}\filldraw[draw=black,fill=unknown] (1,1) rectangle (1.4,1.25); \end{tikzpicture} unknown
    \label{fig:colour_code}}
  \caption[Intermediate steps of our segmentation pipeline]{Intermediate steps of our segmentation pipeline
\subref{fig:pipeline_input} input image
\subref{fig:pipeline_supervoxels} oversegmentation
\subref{fig:pipeline_classifier} conditional label probabilities (here for label \textit{table}, red=0, blue=1)
\subref{fig:pipeline_labeling} final result after MRF.
\subref{fig:colour_code} Colour code of final result}%
  \label{fig:pipeline}
\end{figure}

\begin{figure}[htbp]
  \centering
  \includegraphics[width=0.4\textwidth]{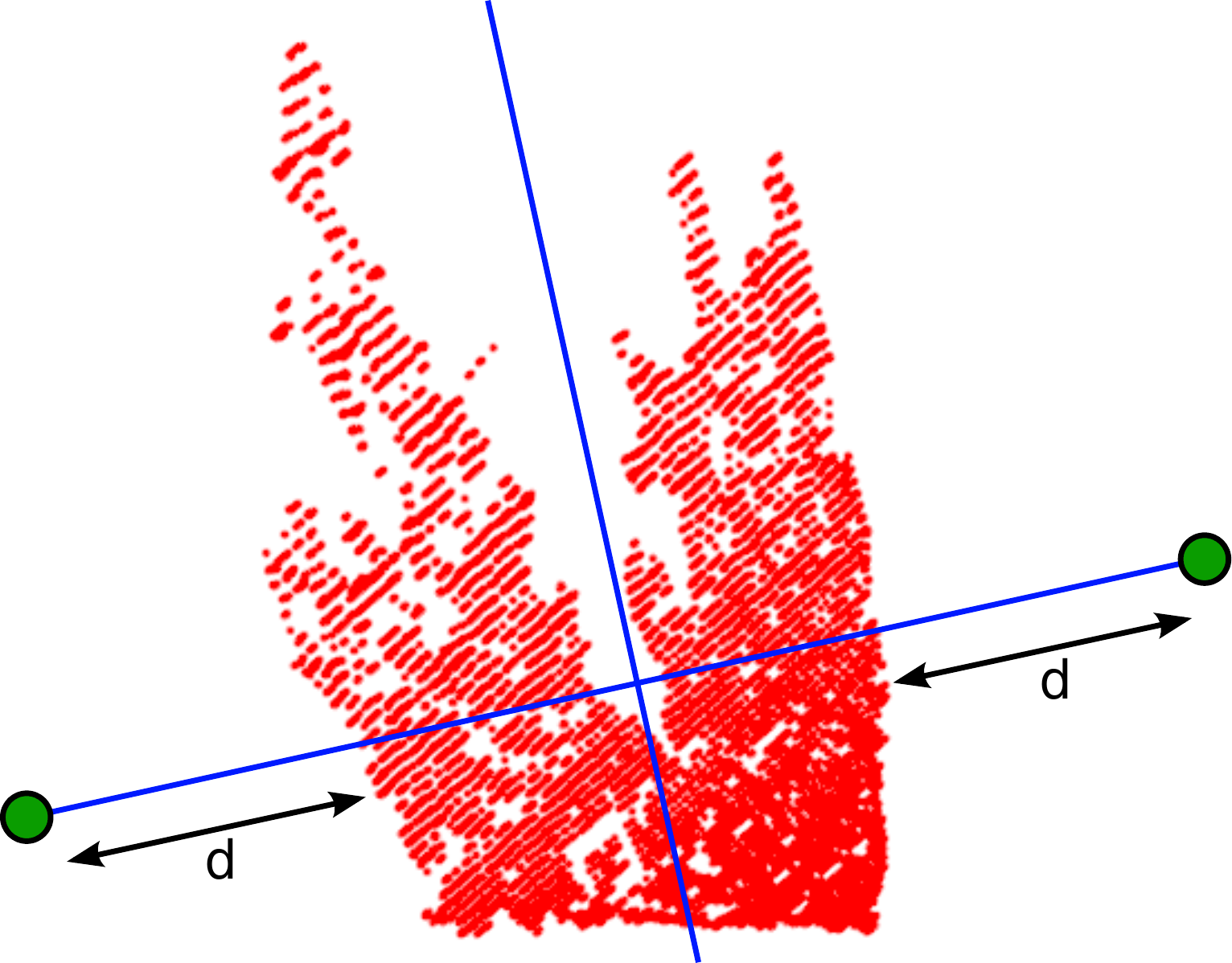}
  \caption{Heuristic to define search positions for a table cluster projected on the ground-plane (red points). Blue lines: Principal axes of the cluster. The search positions (green dots) are placed on the second principal axis, adhering a security distance $d$ from the edge of the table.}
  \label{fig:searchpositions}
\end{figure}

Our proposed point cloud processing pipeline consists of four steps, depicted in Fig. \ref{fig:pipeline}. First, we create an over-segmentation of the scene, clustering it into many small homogeneous patches. In the second step, we compute a manifold but efficient-to-compute feature set for each patch. The resulting feature vector is then processed by a classifier, which yields a probability for each patch being assigned a specific label. To that end, we use a randomized decision forest, a classifier which is intensively discussed in \cite{Criminisi2013}. We train the classifier on the publicly available NYU Depth V2 dataset \cite{Silberman2012}, containing $1{,}449$ indoor frames, recorded by a Microsoft Kinect and densely labelled with more than $1{,}000$ labels. In the last stage of our processing pipeline the classification results set up a pairwise Markov Random Field (MRF), whose optimization yields the final labelling. This last step smoothes the labelling out to correct ambiguous classification results due to noisy local patch information. The final labelling then corresponds to the Maximum-a-posteriori of the output of the MRF. In particular, because our robot can only grasp objects located on tables, we only consider positions next to large clusters of points labelled ``table'' as suitable positions to detect (and consequently grasp) objects. Therefore, after calculating the semantic labels of the current scene, we use simple Euclidean clustering to obtain all tables in the scene. The resulting search positions are then defined by a simple heuristic, which is explained in Fig. \ref{fig:searchpositions}. For further details about our semantic segmentation pipeline we refer to \cite{Wolf2014}.


From this method we gain the knowledge of search locations inside the map, which provide a horizontal plane in a certain height range (to filter out the floor and the lower surface of wall-mounted cupboards or the ceiling). As these locations depend on the first and second principal components derived from the segmented planes we have to filter out positions which are not reachable (e.g. inside of another piece of furniture, or outside of the map) as well as locations behind a wall. For this we first check if the detected table surfaces and the robot are within the same room boundaries and remove any search locations which do not fulfil this criteria. Secondly we do not consider positions that are inside of an occupied grid cell of the static map.

\subsection{Object recognition}
After the robot reaches a ``search position'' an attempt to recognize the object inside the scene and obtain the objects \mbox{6-DOF} pose (position and orientation). The recognizer is called with the RGB-D data collected by the Kinect on the robot's head. It combines three different recognition pipelines (2D and 3D) and merges the results during the verification stage. For details of the recognition process we refer to \cite{aldoma2013multimodal}. Results from successfully recognized objects are the names as well as the poses of all objects. Prerequisite for the recognition process is a trained model of every object which should be localized. 

\subsection{Inform the user}
After the robot recognizes the object at a certain pose the user has to be informed about the current location, therefore the ``search user'' procedure is activated. This procedure will plan a path from the current room to the one in which the user has been detected the last time (either by the robot itself or by a motion detection system), move there and attempt to detect the user by combining 2D face detection technique, introduced by Viola and Jones \cite{viola_jones2004} as well as 3D detection and tracking of human body parts \cite{grammalidis20013, shotton2013real}. If the user is not detected the search will continue until all rooms have been visited by the robot. After detection the user gets informed about the location of objects, during which we use the annotation of the rooms in a reverse order to give the user a pose in human understandable form instead of the 6-DOF pose values.

\section{Experiments}
\label{sec:exp}
\begin{figure}[thpb]
\centering
\includegraphics[scale=1.3]{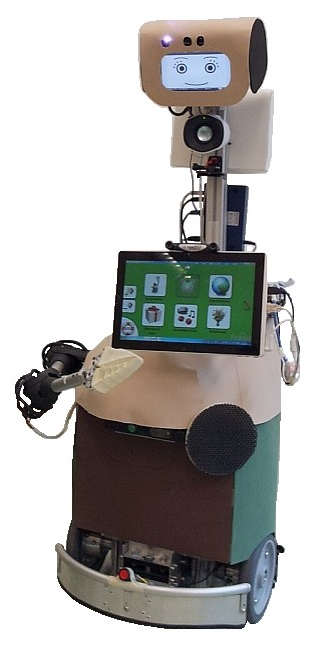}
\caption{HOBBIT platform - prototype 1}
\label{fig:hobbit_pt1}
\end{figure}

All experiments were conducted on the first prototype of our custom made HOBBIT platform \cite{fischingerhobbit} Fig. \ref{fig:hobbit_pt1}. Its base is a mobile platform with two drive motors with an integrated gear and attached wheels with differential drive. The perception system consists of one RGB-D camera (ASUS Xtion Pro Live) mounted ground-parallel at a height of \SI[mode=text]{40}{\centi\metre} and one RGB-D camera (Kinect) mounted at \SI[mode=text]{124}{\centi\metre} height as part of the robot's head on a pan-tilt unit. The lower camera is used for self-localization, the upper one for user and object detection, gesture and object recognition and both cameras contribute to the obstacle detection during navigation. For manipulation PT1 is equipped with an IGUS Robolink arm with a 2-finger Finray-based gripping system. Further parts of the platform include a \US{15}{\inch} tablet computer for touch input and speech recognition as well as a \US{7}{\inch} display for the purpose of displaying the robot's face. To stash away retrieved objects a tray is mounted behind the tablet and above the robot arm.

To evaluate and compare the two implementations described in Section \ref{sec:implementation} we conducted four tests, each consisting of one execution of both implementations. The object that had to be located was placed according to \fref{tab:results}. As objects we used an Asus Xtion Pro box, a brown handbag and an orange wallet, which were placed at the same locations for the corresponding search runs. The robot was placed at the same start location before each pair of runs, with a total of 4 different start locations. 
For each run we collected the following data: 
\begin{itemize}
\item The duration of the entire run until the robot detected the object or ended the search run.
\item The number of ``search positions''. (Pre-defined, calculated and with outliers removed)
\item The success rate of the search run. (Was the object recognized on the actual location?)
\end{itemize}

\begin{table}[h]
\begin{tabular}{llccclll}
\hline
\multicolumn{1}{c}{object} & \multicolumn{1}{c}{start}                              & location                                                         & \multicolumn{2}{c}{\begin{tabular}[c]{@{}c@{}}object\\  detected?\end{tabular}} & \multicolumn{2}{c}{\begin{tabular}[c]{@{}c@{}}duration\\  {[}$\min${]}:{[}$\sec${]}\end{tabular}} & \begin{tabular}[c]{@{}c@{}}\#\end{tabular} \\ \hline
                           &                                                        & \multicolumn{1}{l}{}                                             & \multicolumn{1}{l}{}                   & \multicolumn{1}{l}{}                   &                                      &                                      &                                                                       \\
                           &                                                        & \multicolumn{1}{l}{}                                             & \multicolumn{1}{|c}{P}                 & \multicolumn{1}{|c}{S}                 & \multicolumn{1}{|c}{P}               & \multicolumn{1}{|c}{S}               & \multicolumn{1}{|l}{}                                                 \\ \cline{4-7}
Asus box                   & Kitchen                                                & \begin{tabular}[c]{@{}c@{}}Dining\\ room\\ table\end{tabular}    & Y                                      & Y                                      & 7:12                                 & 16:31                                & 9                                                                     \\ \hline
Handbag                    & \begin{tabular}[c]{@{}c@{}}Living\\ room\end{tabular}  & \begin{tabular}[c]{@{}c@{}}Bedroom\\ nightstand\end{tabular}     & N                                      & N*                                     & 9:52                                 & 12:15                                & 17                                                                    \\ \hline
Wallet                     & \begin{tabular}[c]{@{}c@{}}Central\\ hall\end{tabular} & \begin{tabular}[c]{@{}c@{}}Living\\ room\\ shelf\end{tabular} & Y                                      & N                                      & 7:04                                 & 23:07                                & 36                                                                    \\ \hline
Asus box                   & \begin{tabular}[c]{@{}c@{}}Dining\\ room\end{tabular}  & \begin{tabular}[c]{@{}c@{}}Kitchen\\ cupboard\end{tabular}       & N                                      & Y                                      & 8:51                                 & 10:45                                & 17                                                                   
\end{tabular}
\caption{Test run set-up and results. \#\dots number of ``search positions'', N*\dots false positive recognition of the object, P\dots pre-defined locations, S\dots semantic segmentation-based}
\label{tab:results}
\end{table}

\section{Discussion and outlook}
\begin{figure}[thpb]
\centering
\includegraphics[width=0.4\textwidth]{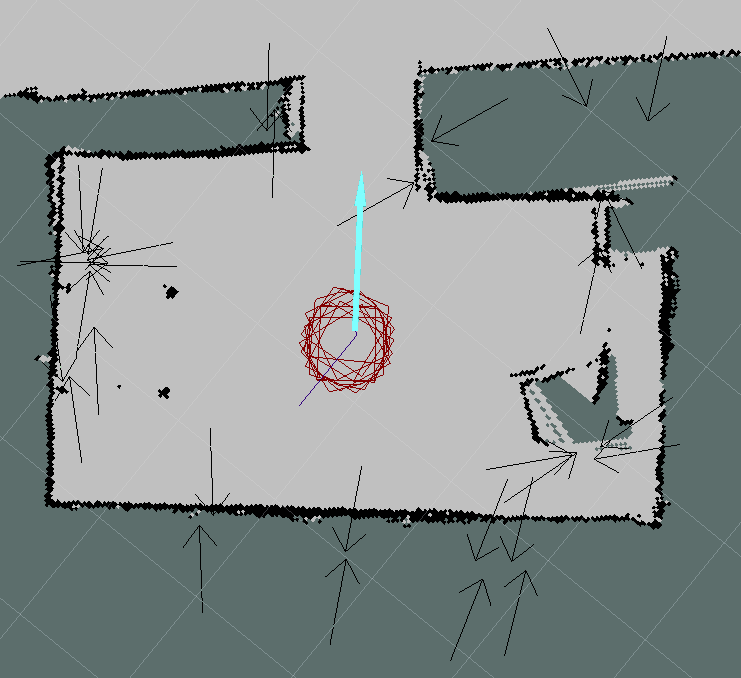} 
\caption{Manual overlay of 12 search position calculations in our lab}
\label{fig:one_room}
\end{figure}

At a first sight the results of our approach uses up to 2.5 times the time of the pre-programmed search task, which was expected due to the fact that the robot not only searches on pre-defined tables but also on window sills, shelves, etcetera.  We further identified 3 major time-costing culprits in our system and set-up. First is the sequential order of the semantic segmentation and the $30\degree$-rotation plus the point cloud acquisition after the entrance of a room. As they both take almost the same time to complete ($5\sec$ and $4\sec$) we can save $48\sec$ for each room. In our 5 virtual room set-up this would lead to a reduction of up to $4\min$, depending on the number of actually visited rooms. This leaves us room for future improvements for our proposed approach. The third culprit is the higher number of ``search positions'' (up to 20) that are inside of the desired room opposed to only one, which leads to a higher probability to find an object at the cost of increased search duration. To reduce this number of positions it is possible to cluster these poses together under the premise that they are close enough and the field of view will overlap at least a certain amount so that a possibly located table plane is not discarded. Fig. \ref{fig:one_room} shows the result of search position calculations without the virtual room constraints and a number of poses which could be clustered together, especially on the left side and in the right handed lower corner.

Another approach is to combine the semantic search algorithm with the pre-defined locations where the ``search locations'' are autonomously generated once instead of the manual labelling. This would improve the speed of the whole search procedure but remove any flexibility towards changes in the environment.

\section{Conclusion}
In this paper we presented a SMACH-based behaviour control for more intelligent indoor fetch-and-carry tasks by the means of semantic segmentation. Our approach does not require manually annotated search positions and thereby increases flexibility and human-oriented perception. We could show the feasibility of the approach by the means of a first series of experiments and outlined suggestions how it can be further improved. After implementing these improvements, we will perform a user study to assess how our approach is evaluated in terms of perceived intelligence and overall satisfaction by naive users. 

\addtolength{\textheight}{-8.5cm}   




\section*{ACKNOWLEDGMENT}
This work has been partially funded by the European Commission under FP7-IST-288146 HOBBIT.

\bibliographystyle{IEEEtran}
\bibliography{root}

\end{document}